\documentclass{article}
\usepackage[T1]{fontenc} % add special characters (e.g., umlaute)
\usepackage[utf8]{inputenc} % set utf-8 as default input encoding
\usepackage{ismir,amsmath,cite,url}
\usepackage{graphicx}
\usepackage{color}
\usepackage{enumitem}
\usepackage{float}
\usepackage{booktabs}
\usepackage{hyperref}

\usepackage{lineno}
\title{Music SketchNet: Controllable Music Generation via Factorized Representations of Pitch and Rhythm}

\multauthor
{Ke Chen$^1$ \hspace{.3cm} Cheng-i Wang$^3$ \hspace{.3cm} Taylor Berg-Kirkpatrick$^2$ \hspace{.3cm} Shlomo Dubnov$^1$ \vspace{.2cm}} { 
$^1$ CREL, Music Department, $^2$ UC San Diego \hspace{.3cm}
$^3$ Smule, Inc\\
{\tt\small$^{1,2}$\{knutchen, tberg, sdubnov\}@ucsd.edu, $^3$cheng-i.wang@smule.com} \vspace{-0.3cm}
}
\def\authorname{Ke Chen,  Cheng-i Wang, Taylor Berg-Kirkpatrick, Shlomo Dubnov}

\begin{document}

\maketitle

\begin{abstract}
Drawing an analogy with automatic image completion systems, we propose Music SketchNet, a neural network framework that allows users to specify partial musical ideas guiding automatic music generation. We focus on generating the missing measures in incomplete monophonic musical pieces, conditioned on surrounding context, and optionally guided by user-specified pitch and rhythm snippets. First, we introduce SketchVAE, a novel variational autoencoder that explicitly factorizes rhythm and pitch contour to form the basis of our proposed model. Then we introduce two discriminative architectures, SketchInpainter and SketchConnector, that in conjunction perform the guided music completion, filling in representations for the missing measures conditioned on surrounding context and user-specified snippets. We evaluate SketchNet on a standard dataset of Irish folk music and compare with models from recent works. When used for music completion, our approach outperforms the state-of-the-art both in terms of objective metrics and subjective listening tests. Finally, we demonstrate that our model can successfully incorporate user-specified snippets during the generation process. 
\end{abstract}

\section{Introduction}
\begin{figure}[t]
 \centerline{
 \includegraphics[width=7.2cm]{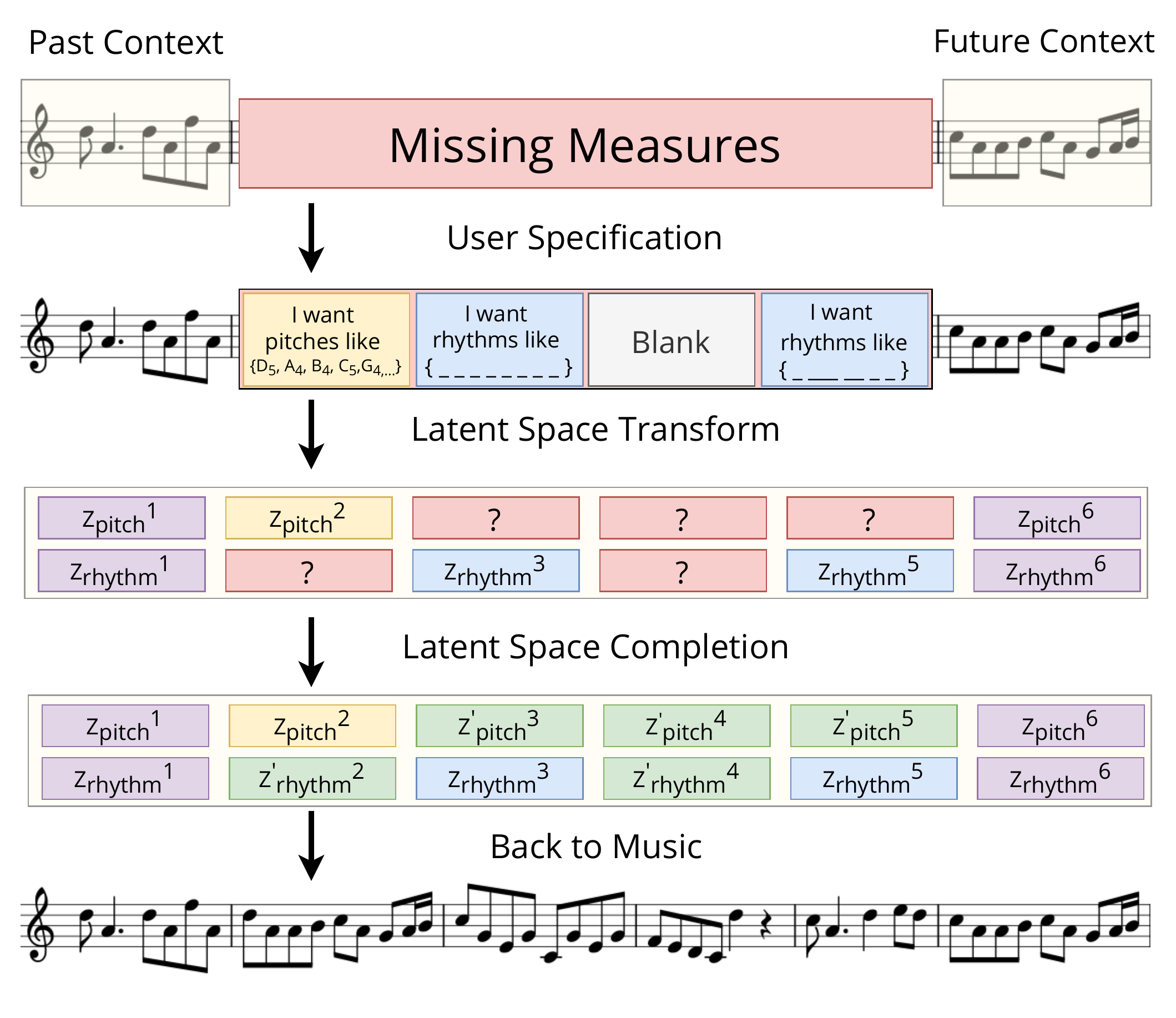}}
 \caption{The music sketch scenario. The model is designed to fill the missing part based on the known context and user's own specification.}
 \label{fig:sketch_scenario}
 \vspace{-0.4cm}
\end{figure}

As a research area, automatic music generation has a long history of studying and expanding human expression/creativity \cite{Loy-composing}. The use of neural network techniques in automatic music generation tasks has shown promising results in recent years \cite{DLTFMG}. In this paper, we focus on a specific facet of the automatic music generation problem on how to allow users to flexibly and intuitively control the outcome of automatic music generation. Prior work supports various forms of conditional music generation. MuseGan \cite{MuseGan} allows users to condition generated results on full-length multi-track music. DeepBach \cite{DeepBach} provides a constraint mechanism that allows users to limit the generated results to match composer styles. Music Transformer \cite{MusicTransformer} supports a accompaniment arrangement from an existing melody track in classical music. However, all these approaches require the user preference to be defined in terms of complete musical tracks. 

Inspired by the sketching and patching work from computer vision \cite{acm-sketch-2d,eccv-sketch-2d,cvpr-sketch-2d,cvpr-sketch-2d-2,sketch-3d}, we propose Music SketchNet\footnote{\href{https://github.com/RetroCirce/Music-SketchNet}{https://github.com/RetroCirce/Music-SketchNet}.}which allows users to specify partial musical ideas in terms of incomplete and distinct pitch and rhythm representations. More specifically, we generalize the concept of sketching and patching -- wherein a user roughly sketches content for a missing portion of an image -- to music, as depicted in Figure \ref{fig:sketch_scenario}. The proposed framework will complete the missing parts given the known context and user input. To the best of our knowledge, there has been limited work on sketching in music generation. Some work \cite{mcmc-music-inpaint,DeepBach} has used Markov Chain Monte Carlo (MCMC) to generate music with given contexts or generate music conditioned on simple starting and ending notes \cite{anticipationrnn}. The most related task is music inpainting: completing a musical piece by generating a sequence of missing measures given the surrounding context, but without conditioning on any form of user preferences. Music InpaintNet,\cite{musicinpaint} completes musical pieces by predicting vector representations for missing measures, then the vector representations are decoded to output symbolic music through the use of a variational autoencoder (VAE) \cite{vae}.

Our proposed music sketching scenario takes music inpainting a step further. We let users specify musical ideas by controlling pitch contours or rhythm patterns, not by complete musical tracks. The user input is optional: users can choose to specify musical ideas, or let the system fill in predictions without conditioning on user preferences. 

\begin{figure}
 \centerline{
 \includegraphics[trim=0 10 10 0, clip, width=\columnwidth]{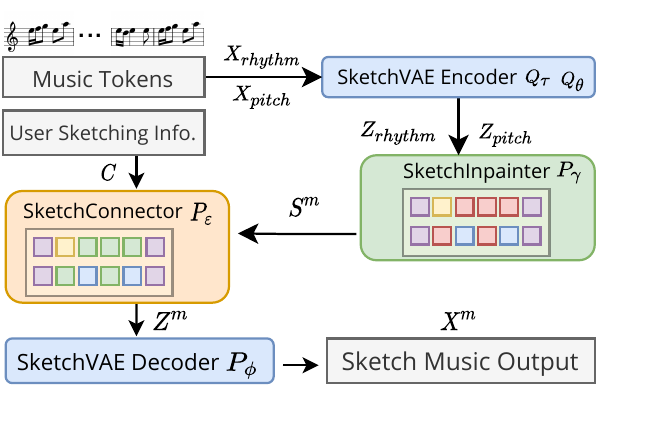}}
 \caption{The Music SketchNet pipeline. The color patterns inside Inpainter and Connector correspond to the latent space transform and completion process in Figure \ref{fig:sketch_scenario}.}
 \label{fig:sketchnet_outline}
\end{figure}

Music SketchNet consists of three component, as depicted in Figure~\ref{fig:sketchnet_outline}: (1) SketchVAE is a novel variational autoencoder that converts music measures into high dimensional latent variables. By the use of a factorized inference network, SketchVAE decouples latent variables into two parts: pitch contour and rhythm, which serve as the control parameters for users. (2) SketchInpainter contains stacked recurrent networks to handle the element-level inpainting prediction in the latent space. (3) SketchConnector receives users' sketches of pitch, rhythm, or both, combines them with the prediction from SketchInpainter, and finalizes the generation.

In this paper, we show that the proposed SketchVAE is capable of factorizing music input into latent variables meaningfully, and the proposed SketchInpainter/SketchConnector allows users to control the generative process. The novel training and evaluation methodologies of the SketchConnector are also presented. 

\section{Music Sketching}
We formalize the music sketching task as solving the following three problems: (1) how to represent music ideas or elements, (2) how to generate new materials given the past and future musical context and (3) how to process users' input and integrate it with the system. A visualization of the sketching scenario is depicted in Figure \ref{fig:sketch_scenario}. 

We propose three neural network components to tackle the three problems. The SketchVAE encodes/decodes the music between external music measures and the learned factorized latent representations. The SketchInpainter predicts musical ideas in the form of the latent variables given known context. And the SketchConnector combines the predictions from SketchInpainter and users' sketching to generate the final latent variables which are sent into the SketchVAE decoder to generate music output. A diagram showing the proposed pipeline is shown in Figure \ref{fig:sketchnet_outline}.   
\subsection{Problem Definition}

More formally, the proposed sketch framework can be described as a joint probability model of the missing musical content, $X^m$, conditioned on the past, future, and user sketching input. The joint probability breaks down into a product of conditional probabilities corresponding to sub-components of the framework: 
\begin{align*}
&P_{\phi,\varepsilon,\gamma,\theta,\tau}(X^{m}, Z, S| X^{p}, X^{f}, C) = \\
&\ P_{\phi}(X^{m} | Z^{m}) \tag{SketchVAE Decoder}\\
& * P_{\varepsilon}(Z^{m} | S^{m}, C) \tag{SketchConnector}\\
& * P_{\gamma}(S^{m}_{pitch} | Z^{p}_{pitch},Z^{f}_{pitch}) \tag{SketchInpainter}\\
& * P_{\gamma}(S^{m}_{rhythm} | Z^{p}_{rhythm},Z^{f}_{rhythm}) \tag{SketchInpainter} \\
& * Q_{\theta}(Z^{p}_{pitch}, Z^{f}_{pitch} | X^{p}_{pitch},X^{f}_{pitch})\\
& * Q_{\tau}(Z^{p}_{rhythm}, Z^{f}_{rhythm} | X^{p}_{rhythm},X^{f}_{rhythm}) \tag{SketchVAE Encoders}
\end{align*}
$X$ indicates the input/output music sequence, $Z$ is the sequence for $\{z\}$ the latent variable
, $S$ is the SketchInpainter's predicted sequence, $C$ is users' sketching input. The superscripts, $p$, $m$, $f$ indicate the past, missing and future context. The subscripts, $pitch$ and $rhythm$ indicate the pitch and rhythm latent variables. $Q_{\theta}$, $Q_{\tau}$, $P_{\phi}$ are the SketchVAE pitch/rhythm encoders and decoder parameters, $P_{\gamma}$ represents the SketchInpainter, and $P_{\varepsilon}$ is the SketchConnector. 

\subsection{SketchVAE for Representation}

MusicVAE \cite{musicvae} is one of the first works applying the variational auto-encoder \cite{vae} to music. MeasureVAE \cite{musicinpaint} further focuses on representing isolated measures and utilizes a hierarchical decoder to handle ticks and beats. $EC^{2}$-VAE \cite{ec2vae} factorizes music measures with separate vectors representing pitch and rhythm by a single encoder and two decoders. Our proposed SketchVAE aims to factorize representations by introducing a factorized encoder that considers pitch and rhythm information separately in the encoder channels. Different from $EC^{2}$-VAE, it could allow users to enter parts of the information (rhythm and/or pitch) optionally.

SketchVAE aims to represent a single music measure as a latent variable $z$ that encodes rhythm and pitch contour information in separate dimensions ($z_{pitch}, z_{rhythm}$). It contains (1) a pitch encoder $Q_{\theta}(z_{pitch} | x_{pitch})$, (2) a rhythm encoder $Q_{\tau}(z_{rhythm} | x_{rhythm})$, and (3) a hierarchical decoder $P_{\phi}(x | z_{pitch}, z_{rhythm})$ as shown in Figure \ref{fig:sketchvae}.

\subsubsection{Music Score Encoding}
\begin{figure}[h]
\centerline{
 \includegraphics[width=\columnwidth]{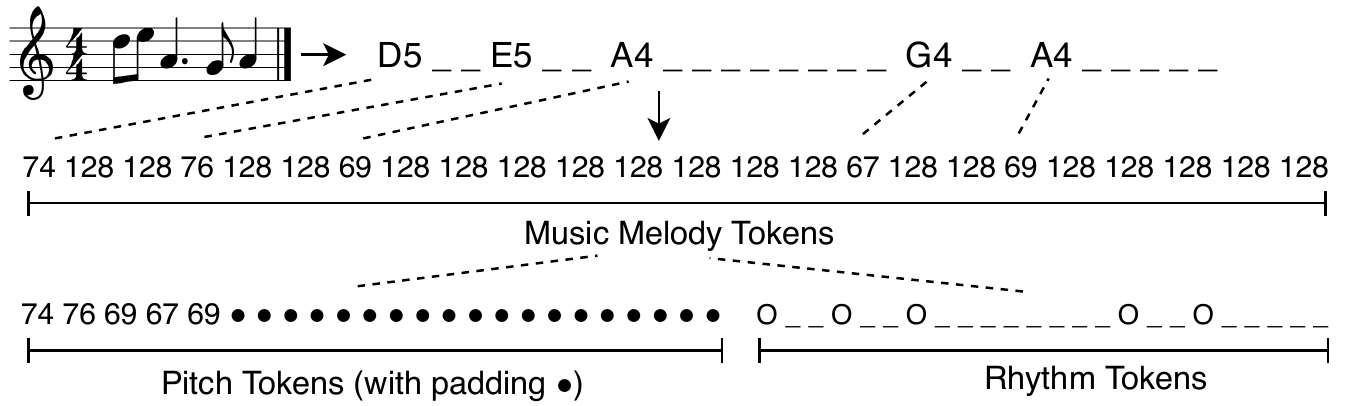}}
 \caption{An example of the encoding of a monophonic melody.}
 \label{fig:musicencoding}
\end{figure}

Similar to \cite{musicvae}, we encode the monophonic midi melody by using [0, 127] for the note onsets, 128 for holding state, and 129 for the rest state. We cut each measure into 24 frames to correctly quantize eighth-note triplets like \cite{musicinpaint}, and encode the midi as described in the previous sentence.

As Figure \ref{fig:musicencoding} shows, we further process the encoded 24-frame sequence $x$ into $x_{pitch}$ and $x_{rhythm}$, the pitch and rhythm token sequences respectively. The pitch token sequence $x_{pitch}$ is obtained by picking all note onsets in $x$ with padding (shown by "$\bullet$" in Figure \ref{fig:musicencoding}) to fill 24 frames. The rhythm token sequence $x_{rhythm}$ is obtained by replacing all pitch onsets with the same token (shown by "O" and "\_" in Figure \ref{fig:musicencoding}). A similar splitting strategy is also used in \cite{mumeAlex}. Our motivation is to provide users with two intuitive music dimensions to control, and to help enforce better factorization in the latent representation for later prediction and control. 

\subsubsection{The Pitch Encoder and Rhythm Encoder}
After pre-processing $x$, $x_{pitch}$ only contains the note value sequence, while $x_{rhythm}$ only has the duration and onset information. $x_{pitch}$ and $x_{rhythm}$ are then fed into two different GRU \cite{GRU} encoders for variational approximation. The outputs of each encoder are concatenated into $z = [z_{pitch}, z_{rhythm}]$.

\subsubsection{The Hierarchical Decoder}
\begin{figure}[t]
 \centerline{
 \includegraphics[width=\columnwidth]{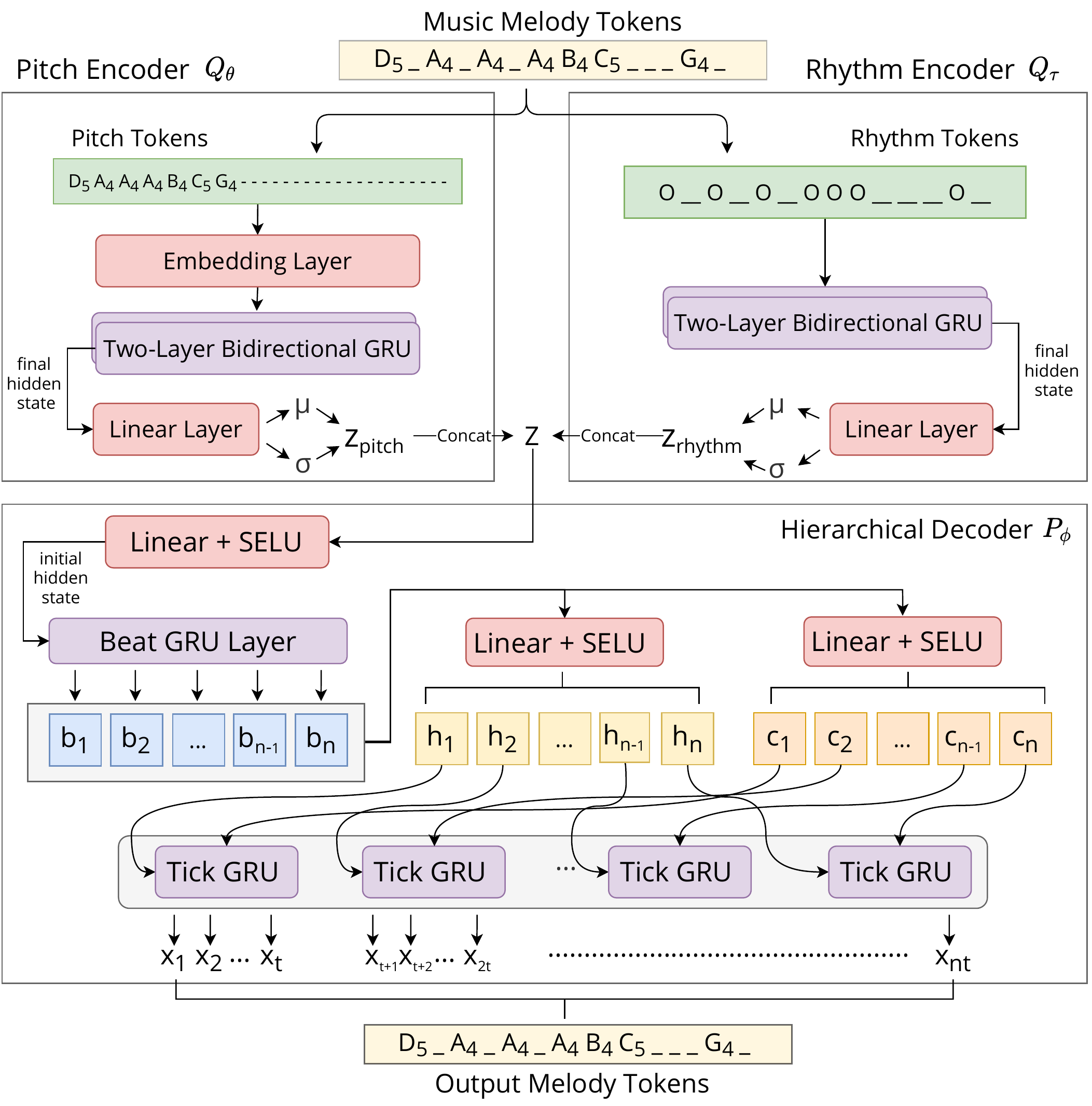}}
 \caption{SketchVAE structure: pitch encoder, rhythm encoder and hierarchical decoder. Rhythm tokens: the upper dashes denote the onsets of note, and the bottom dashes denote the hold/duration state. We use pitch symbols to represent the tokens numbers for better illustration.} 
 \label{fig:sketchvae}
\end{figure}

After we obtain the latent variable $z$, we feed it into the hierarchical decoder. This decoder is similar to the decoder used in MeasureVAE \cite{musicinpaint}. As shown in the bottom part in Figure \ref{fig:sketchvae}, it contains an upper "beat" GRU layer and a lower "tick" GRU layer. This division's motivation is to decode $z$ into $n$ beats first and then decode each beat into $t$ ticks. As a result, the note information in each measure will be decoded in a musically intuitive way. For the tick GRU, we use the teacher forcing \cite{teacherforcing-1,teacherforcing-2} and auto-regressive techniques to train the network efficiently. The output is conditioned frame-by-frame not only on the beat token but also on the last tick token.

% The loss function of the variational autoencoder in this factorization scenario can be described as,
% \begin{align*}
%     L(\theta,\tau,\phi) &= E_{Q_{\theta,\tau}(z|x)}[log P_{\phi}(x|z)] \\
%     & - D_{KL}(Q_{\theta}(z_{pitch}|x) || p(z_{pitch}))\\  
%     & - D_{KL}(Q_{\tau}(z_{rhythm}|x) || p(z_{rhythm})) 
% \end{align*}
% where we set the prior of the latent variables $p(z_{pitch})$ and $p(z_{rhythm})$ to be a Gaussian distribution.

\begin{figure*}[t]
 \centerline{
 \includegraphics[width=16cm]{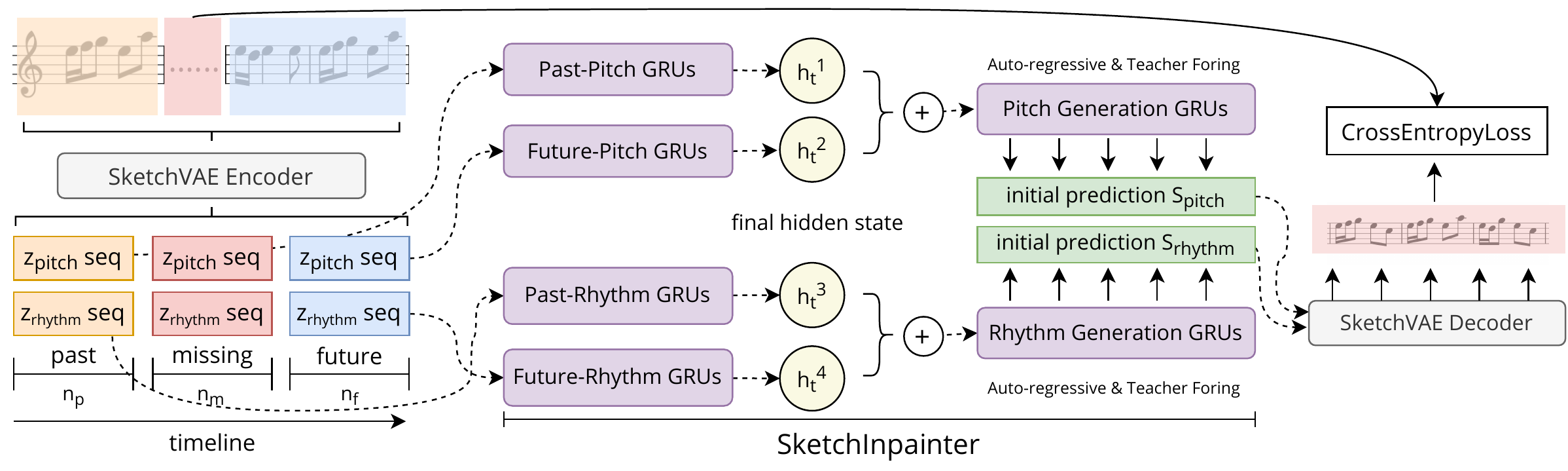}}
 \caption{SketchInpainter structure. We feed the music tokens into the SketchVAE and obtain the latent variable sequences. And we feed the sequences into the pitch GRU and the rhythm GRU groups to generate the initial prediction $S$.}
 \label{fig:sketchgen}
\end{figure*}
\subsubsection{Encoding the Past, Missing and Future Musical Context}
The latent variable sequences $Z^p$, $Z^m$, and $Z^f$ are then obtained by processing the music input in measure sequences $X^p$, ${X^m}$, and ${X^f}$. Both $X^m$ and $Z^m$ are masked during training. This encoding part is shown in the left block of Figure \ref{fig:sketchgen}.

\subsection{SketchInpainter for Initial Prediction}
Next, we describe the model component that performs the music inpainting to predict latent representations for the missing measures. The SketchInpainter accepts $Z_{pitch}$ and $Z_{rhythm}$ as two independent inputs from SketchVAE. Then only the past and future $Z_{pitch}$ and $Z_{rhythm}$ are fed into the pitch/rhythm GRU groups respectively. The output from each GRU group is the hidden state $h$, as shown in the middle of Figure \ref{fig:sketchgen}.

Then we combine the past/future hidden states $h$ from both the pitch and rhythm GRU groups and use them as the initial states for the pitch/rhythm generation GRUs. The generation GRUs then predict the missing latent variables by $S^m = (S_{pitch}, S_{rhythm})$, as shown in the green box in Figure \ref{fig:sketchgen}. Each generation GRU is trained with the teach forcing and auto-regressive techniques.

Each output vector $s^m$ from $S^{m}$ has the same dimension as the latent variable $z$ from $Z$. We first build a model with only SketchVAE and SketchInpainter that directly predicts the missing music material, $X^m$. As the right block of Figure \ref{fig:sketchgen} shows, $S^m$ is sent into the SketchVAE decoder and we compute the cross entropy loss between the predicted music output and the ground truth. This is the stage I training in our model, detailed in Section 3.3.

\subsection{SketchConnector for Finalization}
The predicted $S^m$ from SketchInpainter can already serve as a good latent representation for the missing part $X^m$. We continue by devising the SketchConnector, $P_{\varepsilon}(Z^{m} | S^{m}, C)$, to modify the prediction with user control. To make up for the lack of correlation between pitch and rhythm in current predictions, we introduce the SketchConnector as a way to intervene/control the generative process, that also leads to a wider musical expressivity of the proposed system.

\subsubsection{Random Unmasking}
\begin{figure}[h]
\centerline{
 \includegraphics[width=6.0cm]{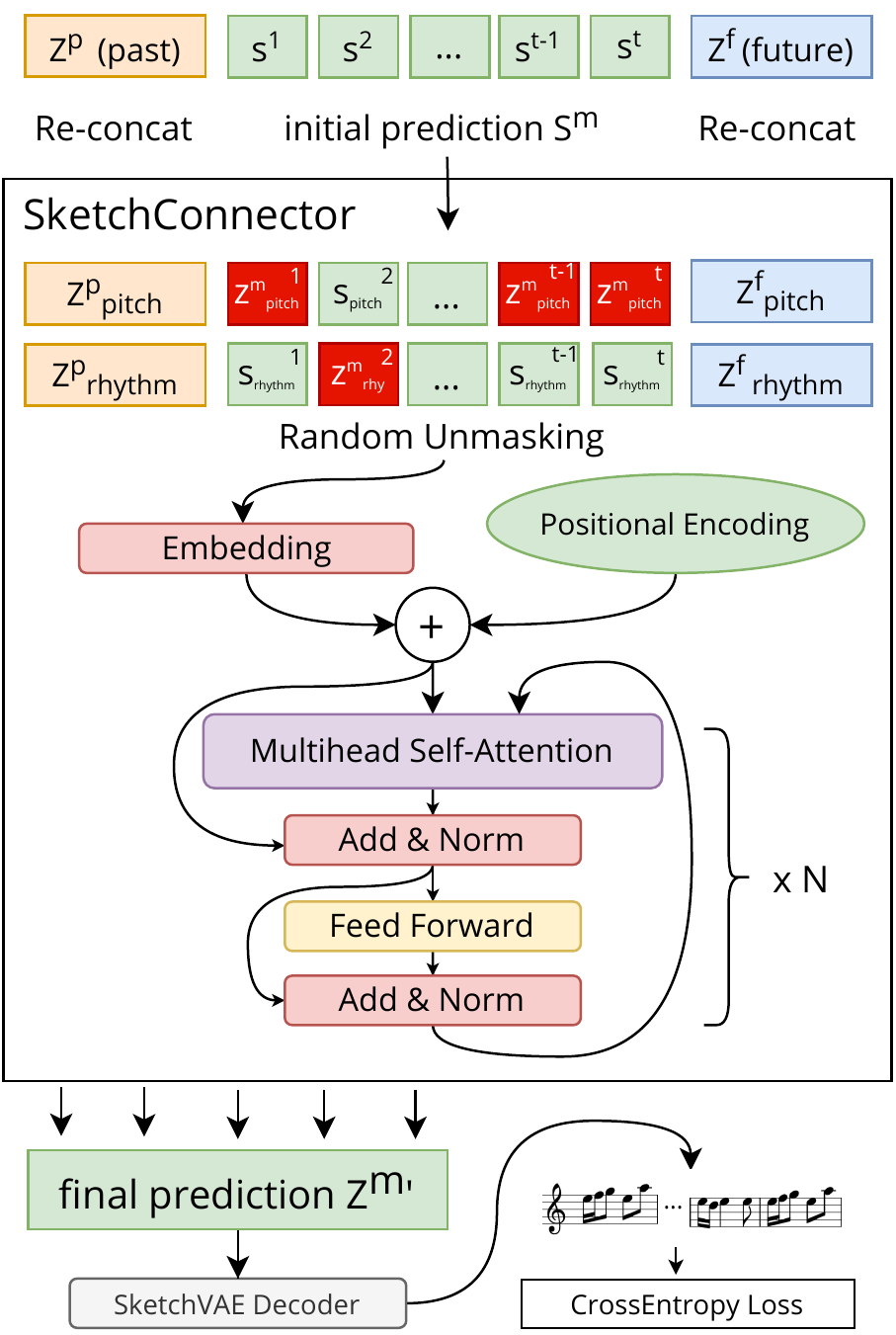}}
 \caption{The SketchConnector: the output of SketchInpainter is randomly unmasked and fed into a transformer encoder to get the final output.}
 \label{fig:sketchconnector}
 \vspace{-0.4cm}
\end{figure}

With $S^m$ obtained from SketchInpainter, we concatenate it with $Z^p$ and $Z^f$ again. However, before we feed it back into the network, we randomly unmask some of the missing parts to be the ground-truth (simulating user providing partial musical context). The masked $S^m$ are shown by the red boxes in Figure \ref{fig:sketchconnector}. We replace some $s$ from $S^m$ to be the real answer in $Z^m$, denoted as $C$. We observe that this optimization is very similar to BERT \cite{BERT} training. The difference is that BERT randomly masks the ground truth labels to be unknown, but SketchConnector randomly unmasks the predictions to be truths. The unmasking rate is set to 0.3.  

Intuitively, this allows the model to learn a more close relation among current rhythm, pitch tokens, and the nearest neighbour tokens. In the sketch inference scenario, the randomly unmasked measures will be replaced by the user sketching information, which allows a natural transition between the training and testing process.

\subsubsection{Transformer-based Connector}

Then with $S^m$ and the random unmasking data $C$, we feed them into a transformer encoder with absolute positional encoding. In contrast to \cite{MusicTransformer}, we do not use relative positional encoding because our inputs are vectors representing individual measures, whose length is far shorter than midi-event sequences. 

\begin{table*}[]
\begin{tabular}{@{}c|ccc|ccc|ccc|@{}}
\toprule
& \multicolumn{3}{c}{Irish-Test} & \multicolumn{3}{c}{Irish-Test-R} & \multicolumn{3}{c}{Irish-Test-NR} \\ \midrule
Model               & loss $\downarrow$     & pAcc $\uparrow$     & rAcc $\uparrow$     & loss $\downarrow$      & pAcc $\uparrow$     & rAcc $\uparrow$     & loss $\downarrow$     & pAcc $\uparrow$      & rAcc $\uparrow$      \\ \midrule
Music InpaintNet            & 0.662    & 0.511    & 0.972    & 0.312     & 0.636     & 0.975    & 0.997     & 0.354     & 0.959     \\ \midrule
SketchVAE + InpaintRNN      & 0.714    & 0.510    & 0.975    & 0.473     & 0.619     & 0.981    & 1.075     & 0.374     & 0.964     \\ \midrule
SketchVAE + SketchInpainter & 0.693    & 0.552    & 0.985    & 0.295     & 0.692     & 0.991    & 1.002     & 0.389     & 0.977     \\ \midrule
SketchNet                   & \textbf{0.516}    & \textbf{0.651}    & \textbf{0.985}    & \textbf{0.206}     & \textbf{0.799}     & \textbf{0.991}    & \textbf{0.783}     & \textbf{0.461}     & \textbf{0.977}     \\ \bottomrule
\end{tabular}
\caption{The generation performance of different models in Irish and Scottish  monophonic music dataset. The InpaintRNN is the generative network in Music InpaintNet.}
\label{tab:sketch-result}
\end{table*}

The output of the SketchConnector, $Z^{m}$, will be the final prediction for the missing part. We feed it into the SketchVAE decoder, and compute the cross entropy loss of the output with the ground-truth.

\section{Experiment}
\subsection{Dataset and Baseline}
To evaluate the SketchVAE independently, we compare our model with two related systems: MeasureVAE \cite{musicinpaint} and $EC^2$-VAE \cite{ec2vae}. For SketchNet, we compare our generation results with Music InpaintNet \cite{musicinpaint}, which has shown better results than the earlier baseline \cite{anticipationrnn}. Similar to \cite{musicinpaint}, we use the Irish and Scottish monophonic music Dataset \cite{IrishDataset} and select the melodies with a 4/4 time signature. About 16000 melodies are used for training and 2000 melodies for testing.

\subsection{SketchVAE Measurements}
\subsubsection{Reconstruction}
For SketchVAE, MeasureVAE and $EC^2$-VAE, the dimension of latent variable $|z|$ is set to 256, half for the pitch contour, and the other half for the rhythm. We set the learning rate to 1e-4 and use Adam Optimization with $\beta_1 = 0.9$ and $\beta_2 = 0.998$. Three models achieve the accuracy (the reconstruction rate of melodies) 98.8\%, 98.7\%, 99.0\% respectively. We can clearly conclude that all VAE models are capable of converting melodies to latent variables by achieving the accuracy around 99\%. SketchVAE is capable of encoding/decoding musical materials in SketchNet.

\subsubsection{Comparison with $EC^2$-VAE}
$EC^{2}$-VAE \cite{ec2vae} is also capable of decoupling the latent variable into rhythm and pitch contour dimensions. However, SketchVAE's encoders can accept pitch contour/rhythm inputs separately. Rhythm and pitch controls can be manipulated independently in the sketching scenario where the user might not specify an entire musical measure (e.g., just a rhythm pattern). By contrast, $EC^{2}$-VAE requires a completed measure before encoding. If users want to specify either rhythm or pitch controls, the model must first fill in the other half part before inputting it, which prohibits the possibility of the separate control.

\begin{figure*}[t]
 \centerline{
 \includegraphics[width=\textwidth]{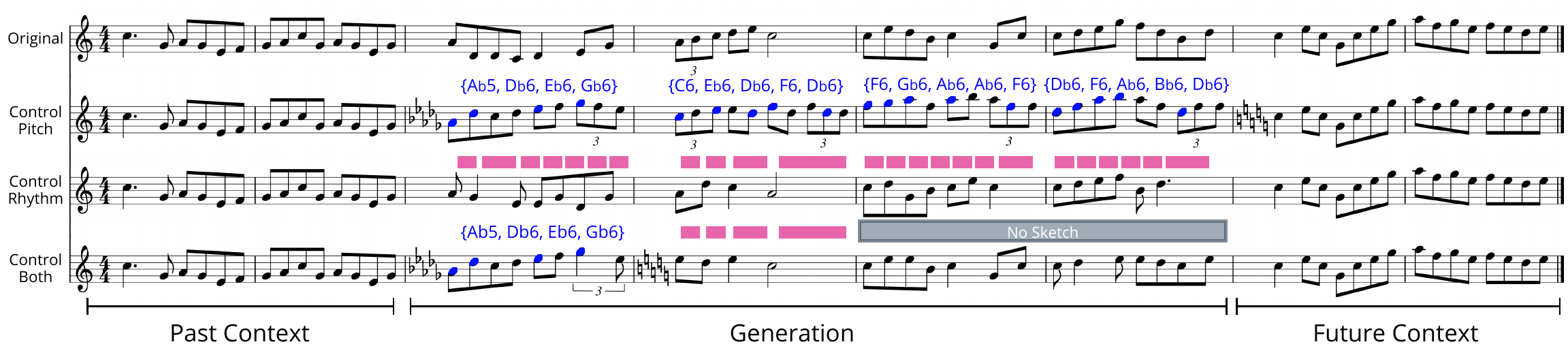}}
 \caption{An example of sketch generation. From top to bottom: original, pitch/rhythm/mixture control. The blue pitch texts denote pitch controls, and the pink segments denote rhythm controls.} 
 \label{fig:sketch_example}
\end{figure*}

\subsection{Generation Performance}
\subsubsection{Training Results}
The SketchNet's training is separated into stage I and II. In stage I, after training the SketchVAE, we freeze its parameters and train the SketchInpainter as shown in the right block of Figure \ref{fig:sketchgen}. In stage II, with the trained SketchVAE and SketchInpainter, we freeze both, concatenate $S^m$ with the past/future latent variables, and feed them to the SketchConnector for training. 

We compare four models by using 6 measures of past and future contexts to predict 4 measures in the middle (i.e. $n_p = n_f = 6$, and $n_m = 4$ ). Music InpaintNet \cite{musicinpaint} is used as the baseline, along with several variations. Early stopping is used for all systems.

\begin{table}[]
\begin{tabular}{@{}cccc@{}}
\toprule
\multicolumn{1}{l}{Model}         & Complexity$\uparrow$ & Structure$\uparrow$ & Musicality$\uparrow$ \\ \midrule
Original                    & 3.22        & 3.47      & 3.56               \\ \midrule
InpaintNet                     & 2.98        & 3.01      & 3.09               \\ \midrule
SketchNet                         & 3.04        & 3.29      & 3.26               \\ \bottomrule
\end{tabular}
\caption{Results of the subjective listening test.}
\label{tab:listen-result}
\end{table}

We compute three metrics: loss, pitch accuracy, and rhythm accuracy to evaluate the model's performance. The pitch accuracy is calculated by comparing only the pitch tokens between each generation and the ground truth (whether the model generates the correct pitch in the correct position). And the rhythm accuracy is calculated by comparing the duration and onset (regardless of what pitches it generates). The overall accuracy and loss are negatively correlated.

For this part of the experiment, we also use two special test subsets. We compute the similarities between the past and future contexts of each song in the Irish test set, pick the top 10\% similar pairs (past and future contexts are almost the same) and bottom 10\% pairs (almost different), and create the Irish-Test-R (repetition) and Irish-Test-NR (non-repetition) subsets.

From Table \ref{tab:sketch-result}, we can see that SketchNet beats all other models for all test sets. The performance improved more for pitch then for rhythm. The accuracy is almost the same between the 1st and 2nd model. Accuracy is slightly better if we use SketchInpainter to treat rhythm and pitch independently during generation. Lastly, with the power of transformer encoder and random unmasking process done in SketchConnector, we can achieve the best performance by using SketchNet (bottom row in Table \ref{tab:sketch-result}). We further follow \cite{sigtest} to use the Bootstrap significance test to verify the difference between each pair's overall accuracy for models in the whole Irish-Test set (Four models, i.e. six pairs in total). The sample time is set to 10000. After calculation, all p-values except the fist and second model pair (p-value = 0.402) are less than 0.05, which proves that SketchNet is different from the left three models.

In the repetition test subsets, the loss of Music InpaintNet is 0.312, which is lower enough to capture repetitions in the musical context and fill in the missing part by copying. In most cases, copying is the correct behaviour because the original melody has repetitive pattern structures. The loss is a measurement to evaluate if the model can learn the repetitive pattern and copy mechanism from the data. the SketchNet slightly outperforms InpaintNet.

% \begin{table}[]
% \centering
% \resizebox{\columnwidth}{!}{%
% \begin{tabular}{cccc}
% \hline
% \small{Pair} & \small{(1) vs (2)} & \small{(1) vs (3)} & \small{(2) vs (3)} \\ \hline
% \small{p-value} & \small{0.01} & \small{0.04} & \small{0.001} \\ \hline
% \end{tabular}%
% }
% \caption{A significance test via ANOVA in the subjective listening test. (1) Original Song (2) Generation by Music InpaintNet (3) Generation by Music SketchNet.}
% \label{tab:sig-test}
% \end{table}
\subsubsection{Subjective Listening Test}
However, the more interesting result is the generation with non-repetition subset. In this case, models cannot merely copy because original melodies do not repeat its content. We see higher losses in all models in this subset compared to the repetition subset. Intuitively, it means that repetitive patterns are essential to the reconstruction task, not necessarily the expressivity of the generated output would be less. 

To further evaluate the proposed SketchNet, we conduct an online subjective listening test to let subjects judge the generated melodies from the non-repetition subset. Each subject will listen to three 32-second piano-rendered melodies: the original, the Music InpaintNet's generation, and the SketchNet's generation. Songs are randomly picked from the Irish-Test-NR set. The beginning and ending (past \& future) are the same for the three melodies. Since the subjective feeling of music is complicated to quantify, we chose three criteria: the number of notes (\textbf{complexity}), the repetitiveness between musical structures (\textbf{structure}), and the degree of harmony of the music (\textbf{overall musicality}). In this way, subjects with different levels of music skills can all give reasonable answers.
% \begin{itemize}[leftmargin = 10pt]
%     \item Complexity: Are the notes complex enough, or just 1-2 single notes?
%     \item Structure: How well does the generated melody share consistency with the past and future music materials?
%     \item Musicality: The overall musicality from the subject.
% \end{itemize}

Before rating the songs, subjects will see three criteria descriptions as we introduced below. The rating is ranged from 1.0 to 5.0 with a 0.5 step. We collected 318 surveyed results from 106 subjects (each subject listens to three groups, nine melodies in total). The average rating of each criteria for all models are shown in Table \ref{tab:listen-result}. The subjective evaluations of all three criteria in SketchNet are better for those of Music InpaintNet. Similar to section 3.3.1, we also conduct a pairwise significance test via Bootstrap in three criteria. All p-values except the <complexity: InpaintNet, SketchNet> (p-value = 0.364) are less than 0.05. It proves that three models (including original songs) are significantly different in structure and overall musicality (subjective feeling to a person). As for the complexity, we believe that the results generated by the two models are similar in terms of the richness of notes, and our model does not significantly increase the number of notes generated.

% With the results above, we can see that SketchNet achieve an overall better performance than the previous models. In particular, from the perspective of reconstruction, SketchNet achieves the lowest loss (0.206) and highest pitch and rhythm accuracy in the repetition subset. Moreover, from the perspective of generation, we observe better subjective ratings for SketchNet than the ratings for Music InpaintNet in the non-repetition subset. Though both models still have room to improve in comparison to the original music.
\begin{table}[]
\centering
\resizebox{5.5cm}{!}{%
\begin{tabular}{cllclllclll}
\hline
\multicolumn{3}{c}{\small{Control Info.}} & \multicolumn{4}{c}{\small{Rhythm}} & \multicolumn{4}{c}{\small{Pitch}} \\ \hline
\multicolumn{3}{c}{\small{Pitch Acc.}} & \multicolumn{4}{c}{\small{0.189}} & \multicolumn{4}{c}{\small{\textbf{0.881}}} \\
\multicolumn{3}{c}{\small{Rhythm Acc.}} & \multicolumn{4}{c}{\small{\textbf{0.973}}} & \multicolumn{4}{c}{\small{0.848}} \\ \hline
\end{tabular}%
}
\caption{The accuracy of the virtual control experiment.}
\label{tab:v-test}
\end{table}

\subsection{Sketch Scenario Usage}
The contribution of Music SketchNet is not only shown in the performance of the generation in section 3.3, but can also be shown in the interactive scenario where users can control the generated output by specifying the rhythm or pitch contour in each measure.

Figure \ref{fig:sketch_example} shows an example of a non-repetition subset melody, where the first and last two measures are given, and the middle parts is generated. The first track is the original melody, the second track is generated with the pitch contour control, the third track is generated with the rhythm control, and the fourth track is controlled with both pitch and rhythm. We can see that each generated melody follows the control from users and develops music phrases accordingly in the missing part. Moreover, each measure is in line with the past and future measures even in the case of scale shift. 

We also provide a "virtual control experiment" to statistically show that users' control did influence the model's generation process. We randomly collect 3000 sample pairs (A, B) from the Irish-Test set. And we use the pitch/rhythm of Sample B to be the sketch information in the same missing position of Sample A. Then we let the model make the generation. We then compute the pitch/rhythm accuracy\footnote{The metric to calculate the pitch accuracy is different from section 3.3.1, because the generated pitches in the new song might have different onset positions. We leverage the Longest Common Sequence to calculate the accuracy. The implementation is presented in the code archive.} in the missing position between the generation and Song B. From \ref{tab:v-test} we can see if we sketch song B's rhythm into the model, the generation will follow the rhythm with 97.3\% accuracy but has different (18.9\%) pitches. However, when we sketch pitches, the pitches in the generation will be highly (88.1\%) in line with the sketching. This proves that the user's control has a relatively high guiding effect on the result of the model generated at the specified position.

\section{Conclusion \& Future Work}
In this paper, we propose a new framework to explore decoupling latent variables in music generation. We further convert this decoupling into controllable parameters that can be specified by the user. The proposed Music SketchNet achieves the best results in the objective and subjective evaluations. Practically, we show the framework's application for the music sketching scenario where users can control the pitch contour and/or rhythm of the generated results. There are several possible extensions for this work. Music elements other than pitch and rhythm can be applied into the music sketching scenario by the latent variable decoupling. Also, how to represent a polyphonic music piece in the latent space is another pressing issue. Both are future works that can generalize this model to more applied scenarios.

\section{Acknowledgement}

We would like to thank Cygames for the partial support of this research. 

% For bibtex users:
\bibliography{ISMIR2020template}

\begin{thebibliography}{10}
\providecommand{\url}[1]{#1}
\csname url@samestyle\endcsname
\providecommand{\newblock}{\relax}
\providecommand{\bibinfo}[2]{#2}
\providecommand{\BIBentrySTDinterwordspacing}{\spaceskip=0pt\relax}
\providecommand{\BIBentryALTinterwordstretchfactor}{4}
\providecommand{\BIBentryALTinterwordspacing}{\spaceskip=\fontdimen2\font plus
\BIBentryALTinterwordstretchfactor\fontdimen3\font minus
  \fontdimen4\font\relax}
\providecommand{\BIBforeignlanguage}[2]{{%
\expandafter\ifx\csname l@#1\endcsname\relax
\typeout{** WARNING: IEEEtran.bst: No hyphenation pattern has been}%
\typeout{** loaded for the language `#1'. Using the pattern for}%
\typeout{** the default language instead.}%
\else
\language=\csname l@#1\endcsname
\fi
#2}}
\providecommand{\BIBdecl}{\relax}
\BIBdecl

\bibitem{Loy-composing}
G.~Loy, \emph{Composing with Computers - a Survey of Some Compositional
  Formalisms and Music Programming Languages}.\hskip 1em plus 0.5em minus
  0.4em\relax MIT Press, 1990.

\bibitem{DLTFMG}
J.~Briot, G.~Hadjeres, and F.~Pachet, \emph{Deep Learning Techniques for Music
  Generation}.\hskip 1em plus 0.5em minus 0.4em\relax Springer, 2020.

\bibitem{MuseGan}
H.~Dong, W.~Hsiao, L.~Yang, and Y.~Yang, ``Musegan: Multi-track sequential
  generative adversarial networks for symbolic music generation and
  accompaniment,'' in \emph{Proceedings of the Thirty-Second {AAAI} Conference
  on Artificial Intelligence}.\hskip 1em plus 0.5em minus 0.4em\relax {AAAI}
  Press, 2018, pp. 34--41.

\bibitem{DeepBach}
G.~Hadjeres, F.~Pachet, and F.~Nielsen, ``Deepbach: a steerable model for bach
  chorales generation,'' in \emph{Proceedings of the 34th International
  Conference on Machine Learning, {ICML}}, 2017, pp. 1362--1371.

\bibitem{MusicTransformer}
C.~A. Huang, A.~Vaswani, J.~Uszkoreit, I.~Simon, C.~Hawthorne, N.~Shazeer,
  A.~M. Dai, M.~D. Hoffman, M.~Dinculescu, and D.~Eck, ``Music transformer:
  Generating music with long-term structure,'' in \emph{7th International
  Conference on Learning Representations, {ICLR}}, New Orleans, LA, USA.

\bibitem{acm-sketch-2d}
C.~Barnes, E.~Shechtman, A.~Finkelstein, and D.~B. Goldman, ``Patchmatch: a
  randomized correspondence algorithm for structural image editing,''
  \emph{{ACM} Trans. Graph.}, vol.~28, no.~3, p.~24, 2009.

\bibitem{eccv-sketch-2d}
Y.~G{\"{u}}{\c{c}}l{\"{u}}t{\"{u}}rk, U.~G{\"{u}}{\c{c}}l{\"{u}}, R.~van Lier,
  and M.~A.~J. van Gerven, ``Convolutional sketch inversion,'' in
  \emph{Computer Vision {ECCV} Workshops}.\hskip 1em plus 0.5em minus
  0.4em\relax Amsterdam, The Netherlands: Springer, 2016, pp. 810--824.

\bibitem{cvpr-sketch-2d}
P.~Sangkloy, J.~Lu, C.~Fang, F.~Yu, and J.~Hays, ``Scribbler: Controlling deep
  image synthesis with sketch and color,'' in \emph{2017 {IEEE} Conference on
  Computer Vision and Pattern Recognition, {CVPR}}.\hskip 1em plus 0.5em minus
  0.4em\relax Honolulu, HI, USA: {IEEE} Computer Society, 2017, pp. 6836--6845.

\bibitem{cvpr-sketch-2d-2}
Q.~Yu, F.~Liu, Y.~Song, T.~Xiang, T.~M. Hospedales, and C.~C. Loy, ``Sketch me
  that shoe,'' in \emph{2016 {IEEE} Conference on Computer Vision and Pattern
  Recognition, {CVPR}}.\hskip 1em plus 0.5em minus 0.4em\relax Las Vegas, NV,
  USA: {IEEE} Computer Society, 2016, pp. 799--807.

\bibitem{sketch-3d}
K.~Xu, K.~Chen, H.~Fu, W.~Sun, and S.~Hu, ``Sketch2scene: sketch-based
  co-retrieval and co-placement of 3d models,'' \emph{{ACM} Trans. Graph.},
  2013.

\bibitem{mcmc-music-inpaint}
J.~Sakellariou, , F.~Tria, L.~Vittorio, and F.~Pachet, ``Maximum entropy model
  for melodic patterns,'' in \emph{ICML Workshop on Constructive Machine
  Learning}, 2015.

\bibitem{anticipationrnn}
G.~Hadjeres and F.~Nielsen, ``Anticipation-rnn: enforcing unary constraints in
  sequence generation, with application to interactive music generation,''
  \emph{Neural Computing and Applications}, 2018.

\bibitem{musicinpaint}
A.~Pati, A.~Lerch, and G.~Hadjeres, ``Learning to traverse latent spaces for
  musical score inpainting,'' in \emph{Proceedings of the 20th International
  Society for Music Information Retrieval Conference, {ISMIR}}, Delft, The
  Netherlands, 2019, pp. 343--351.

\bibitem{vae}
D.~P. Kingma and M.~Welling, ``Auto-encoding variational bayes,'' in \emph{2nd
  International Conference on Learning Representations, {ICLR}}, Banff, AB,
  Canada, 2014.

\bibitem{musicvae}
A.~Roberts, J.~H. Engel, C.~Raffel, C.~Hawthorne, and D.~Eck, ``A hierarchical
  latent vector model for learning long-term structure in music,'' in
  \emph{Proceedings of the 35th International Conference on Machine Learning,
  {ICML}}.\hskip 1em plus 0.5em minus 0.4em\relax Stockholm, Sweden: {PMLR},
  2018, pp. 4361--4370.

\bibitem{ec2vae}
R.~Yang, D.~Wang, Z.~Wang, T.~Chen, J.~Jiang, and G.~Xia, ``Deep music analogy
  via latent representation disentanglement,'' in \emph{Proceedings of the 20th
  International Society for Music Information Retrieval Conference, {ISMIR}},
  Delft, The Netherlands, 2019, pp. 596--603.

\bibitem{mumeAlex}
B.~Genchel, A.~Pati, and A.~Lerch, ``Explicitly conditioned melody generation:
  {A} case study with interdependent rnns,'' in \emph{Proceedings of the 7th
  International Workshop on Musical Meta-creation, {MUME}}, 2019.

\bibitem{GRU}
K.~Cho, B.~van Merrienboer, {\c{C}}.~G{\"{u}}l{\c{c}}ehre, D.~Bahdanau,
  F.~Bougares, H.~Schwenk, and Y.~Bengio, ``Learning phrase representations
  using {RNN} encoder-decoder for statistical machine translation,'' in
  \emph{Proceedings of the 2014 Conference on Empirical Methods in Natural
  Language Processing, {EMNLP}}.\hskip 1em plus 0.5em minus 0.4em\relax Doha,
  Qatar: {ACL}, 2014, pp. 1724--1734.

\bibitem{teacherforcing-1}
S.~Bengio, O.~Vinyals, N.~Jaitly, and N.~Shazeer, ``Scheduled sampling for
  sequence prediction with recurrent neural networks,'' in \emph{Advances in
  Neural Information Processing Systems 28: Annual Conference on Neural
  Information Processing Systems}, Montreal, Quebec, Canada, 2015, pp.
  1171--1179.

\bibitem{teacherforcing-2}
A.~Goyal, A.~Lamb, Y.~Zhang, S.~Zhang, A.~C. Courville, and Y.~Bengio,
  ``Professor forcing: {A} new algorithm for training recurrent networks,'' in
  \emph{Advances in Neural Information Processing Systems 29: Annual Conference
  on Neural Information Processing Systems}, Barcelona, Spain, 2016, pp.
  4601--4609.

\bibitem{BERT}
J.~Devlin, M.~Chang, K.~Lee, and K.~Toutanova, ``{BERT:} pre-training of deep
  bidirectional transformers for language understanding,'' in \emph{Proceedings
  of the 2019 Conference of the North American Chapter of the Association for
  Computational Linguistics: Human Language Technologies, {NAACL-HLT}}.\hskip
  1em plus 0.5em minus 0.4em\relax Minneapolis, MN, USA: Association for
  Computational Linguistics, 2019, pp. 4171--4186.

\bibitem{IrishDataset}
B.~L. Sturm, J.~F. Santos, O.~Ben{-}Tal, and I.~Korshunova, ``Music
  transcription modelling and composition using deep learning,'' in
  \emph{Conference on Computer Simulation of Musical Creativity, {CSMC}}, 2016.

\bibitem{sigtest}
T.~Berg{-}Kirkpatrick, D.~Burkett, and D.~Klein, ``An empirical investigation
  of statistical significance in {NLP},'' in \emph{Proceedings of the 2012
  Joint Conference on Empirical Methods in Natural Language Processing and
  Computational Natural Language Learning, {EMNLP-CoNLL}}.\hskip 1em plus 0.5em
  minus 0.4em\relax {ACL}, 2012, pp. 995--1005.

\end{thebibliography}

\end{document}